\documentclass[11pt]{article}

\usepackage[preprint]{acl}
\usepackage{times}
\usepackage{latexsym}
\usepackage[T1]{fontenc}
\usepackage[utf8]{inputenc}
\usepackage{microtype}
\usepackage{inconsolata}
\newcommand{\BaseVideoCount}{899}

\newcommand{\ExtensionVideoCount}{126}

\newcommand{\StandaloneVideoCount}{1,025}
\newcommand{\StandaloneQACount}{4,388}

\newcommand{\StandaloneMinDurationMinutes}{19.93}

\newcommand{\StandaloneMaxDurationHours}{3.86}

\newcommand{\TVEpisodeCount}{402}
\newcommand{\TVSeriesCount}{2}
\newcommand{\TVSeasonCount}{17}
\newcommand{\TVHours}{193.299}
\newcommand{\TVQACount}{826}

\newcommand{\TVSeriesTextCount}{36}
\newcommand{\TVSeriesSpatiotemporalCount}{24}

\newcommand{\TVMaximumBoundEpisodes}{278}
\newcommand{\TVMaximumBoundHours}{97.98}
\newcommand{\TVPrunedQACount}{15}

\newcommand{\TotalQACount}{5,214}
\newcommand{\TotalHours}{887.480}
\newcommand{\RoundedTotalHours}{887}

\newcommand{\MemoryRetrievalTopK}{12}
\newcommand{\MemoryNeighborWindow}{1}

\newcommand{\RetainedQAMinimum}{3}
\newcommand{\RetainedQAMaximum}{5}
\newcommand{\EvidenceClipMinimum}{2}
\newcommand{\EvidenceClipMaximum}{4}

\newcommand{\EvaluationSnapshotQACount}{5,214}
\newcommand{\EvaluationSnapshotTVQACount}{826}

\newcommand{\EvaluationSnapshotDate}{September 25, 2026}
\newcommand{\EvaluationSnapshotTime}{08:36 UTC}
\newcommand{\EvaluationSnapshotModelCount}{13}

\usepackage{amsmath}
\usepackage{amssymb}
\usepackage{subcaption}
\usepackage{array}
\usepackage{pifont}
\usepackage{booktabs}
\usepackage{graphicx}
\usepackage{xcolor}
\usepackage{url}
\usepackage[capitalize,noabbrev]{cleveref}
\usepackage{tikz}
\usetikzlibrary{arrows.meta,positioning,fit}

\definecolor{fourblue}{HTML}{0072B2}
\definecolor{fourorange}{HTML}{E69F00}
\definecolor{fourgreen}{HTML}{009E73}
\definecolor{foursky}{HTML}{56B4E9}
\definecolor{fourvermillion}{HTML}{D55E00}
\definecolor{fourgray}{HTML}{666666}
\definecolor{fourlight}{HTML}{F4F6F8}

\hypersetup{
  colorlinks=true,
  citecolor=fourblue,
  linkcolor=fourblue,
  urlcolor=fourblue
}

\newcommand{\method}{\textbf{4DMemory Agent}}
\newcommand{\dataset}{\textbf{LongVideo}}
\newcommand{\benchmark}{\textbf{4DMemoryBench}}
\newcommand{\harness}{\textbf{4DMemory Harness}}

\newcommand{\cmark}{\textcolor{green}{\ding{51}}}
\newcommand{\xmark}{\textcolor{red}{\ding{55}}}

\AddToHookNext{shipout/foreground}{%
  \begin{tikzpicture}[remember picture,overlay]
    \node[anchor=north west,inner sep=0pt]
      at ([xshift=2.5cm,yshift=-0.6cm]current page.north west)
      {\includegraphics[height=0.9cm]{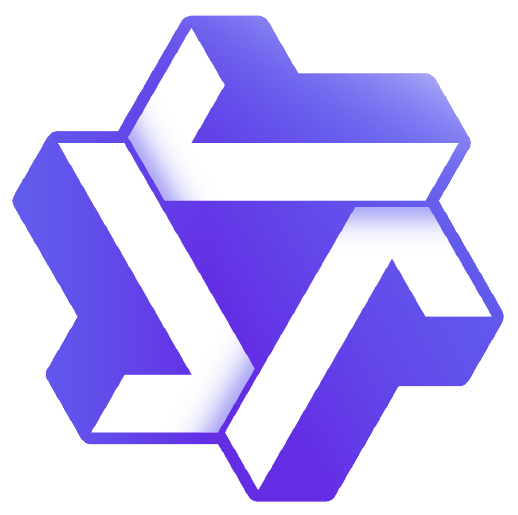}};

    \draw[black,line width=0.4pt]
      ([xshift=2.5cm,yshift=-1.8cm]current page.north west)
      --
      ([xshift=-2.5cm,yshift=-1.8cm]current page.north east);
  \end{tikzpicture}%
}

\title{4DMemory: Benchmarking Long-Horizon Spatiotemporal Memory \\with Harness-Augmented Agents}

\author{
  \setcounter{footnote}{1}%
  \textbf{Yanhao Jia\textsuperscript{1}},
  \textbf{Yi Zhu\textsuperscript{2}\thanks{Corresponding authors.}},
  \textbf{Yiran Zhong\textsuperscript{2}\footnotemark[2]},
  \textbf{Weigao Sun\textsuperscript{2}},
  \textbf{Feida Zhu\textsuperscript{2}}, \\
  \textbf{Pengkun Jiao\textsuperscript{2}},
  \textbf{Sihan Yang\textsuperscript{3}},
  \textbf{Ming Ma\textsuperscript{4,5}},
  \textbf{Steven Hoi\textsuperscript{2}},
  \textbf{Erik Cambria\textsuperscript{1}} \\
  \textsuperscript{1} Nanyang Technological University
  \textsuperscript{2} Token Foundry, Alibaba \\
  \textsuperscript{3} The Chinese University of Hong Kong \\
  \textsuperscript{4} Institute of Neuroscience, Chinese Academy of Sciences \\
  \textsuperscript{5} University of Chinese Academy of Sciences
}

\begin{document}

\maketitle

\begin{abstract}
Humans efficiently process multimodal streams spanning hours to years, building persistent, fine-grained spatiotemporal memories that support reasoning about and predicting long-horizon events. In contrast, existing benchmarks focus on short- and medium-context tasks, and most foundation models struggle to preserve fine-grained information over long horizons. To evaluate and improve long-horizon spatiotemporal reasoning, we introduce the \textbf{LongVideo} dataset and the \textbf{4DMemoryBench}. LongVideo comprises about 900 hours of video and over 5,000 manually reviewed questions covering visual, textual, and spatiotemporal reasoning across diverse context lengths. We further develop the \textbf{4DMemory Harness}, which integrates four complementary skills, 22+ specialized tool models, and a model-specific retrieval-augmented generation system. Evaluating 13 leading models, we find that standalone foundation models struggle with long-horizon tasks, while the 4DMemory Harness improves all of them by over 50\%. Compared to previous benchmark, the same models exhibit a performance drop of more than 10\% on 4DMemoryBench. We hope 4DMemory will facilitate the development of spatiotemporal agents capable of addressing complex, long-horizon tasks.
\end{abstract}

% 人类能够在理解问题后高效地解析从数小时到数年的连续多模态信息流，构建超长的时空记忆以此为基础进行长程复杂问题的推理和预测。然而，目前的基准测试普遍测试短程或中程任务，且大部分基础模型的上下文长度无法完整保留细粒度的长程记忆。为了评价并增强基础模型在长期记忆上的时空推理能力，我们构建了LongVideo Dataset并基于此构建了4DMemory Benchmark，这是一个包含约900小时视频和5000+道问题的数据集，经过人工审核后涵盖了不同上下文长度的视觉、文本和时空问题。此外，我们构建了4DMemory Harness来辅助增强现有模型的推理能力，它包含了四种不同的skills，22种不同的工具模型和model-specifical RAG系统。我们在4DMemory Benchmark上对现有的12个先进模型进行了实验，结果表明单独的基础模型无法完成复杂任务。在4DMemory Harness的辅助下，所有模型的能力都得到了60%以上的性能提升。我们希望4DMemory Benchmark和4DMemory Harness可以帮助社区更高效地构建时空智能体并处理长程复杂任务。

\section{Introduction}
\label{sec:introduction}
\vspace{-2.5mm}

Humans do not always know in advance which aspects of an unfolding experience will matter to a future goal. They form long-term memories as the world evolves and, once a concrete goal arises, selectively retrieve relevant experiences, integrate them with new observations, and update their current understanding~\citep{tulving1973encoding, hupbach2007reconsolidation}. This interaction between persistent memory and goal-directed recall enables reasoning across long temporal horizons. As multimodal large language models (MLLMs) advance from static-image perception toward video understanding and multimodal reasoning~\citep{maaz2023videochatgpt, li2024llavaonevision, team2024gemini, team2024qwen2}, developing an analogous capability becomes increasingly important for artificial systems. Long-horizon 4D tasks require models to preserve how entities, spatial relations, states, and events evolve over time, so that observations made hours earlier can inform later reasoning.

\begin{figure*}[!t]
    \centering
    \includegraphics[width=\linewidth]{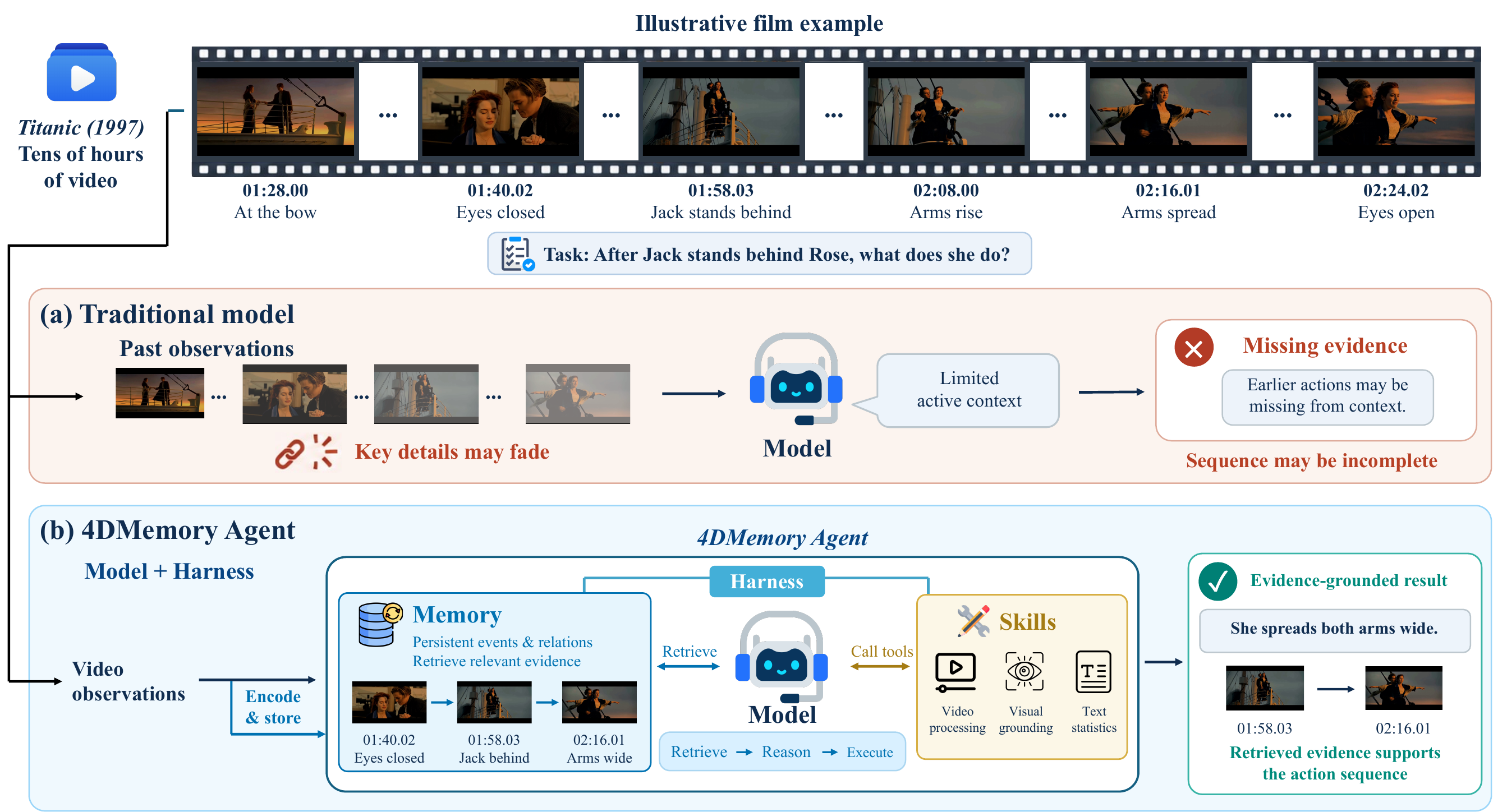}
    \vspace{-8mm}
    \caption{Motivation for 4DMemory Agent. (a) Traditional models with limited active context may lose access to earlier observations, leaving relevant evidence unavailable for temporal reasoning. (b) 4DMemory Agent coordinates persistent memory and task-specific skills through a harness to retrieve evidence and reason over events and their relations.}
    \vspace{-8mm}
    \label{fig:motivation}
\end{figure*}

Although modern MLLMs perform strongly on many vision-language tasks, they remain poorly matched to long-horizon 4D reasoning. First, a nominally long context window does not imply an equally long effective video context, as dense video generates long sequences of visual tokens that impose substantial computational and memory costs during visual encoding and attention. Practical systems therefore depend on sparse frame sampling or token compression, which may discard fine-grained evidence before the eventual task is known~\citep{wu2024longvideobench, fu2025videomme}. Second, strong long-text reasoning does not by itself provide the multimodal alignment, temporal reasoning, and spatial understanding needed to connect events across changing scenes~\citep{liu2024tempcompass, yang2025vsibench}. As a result, directly processing hours of video with an MLLM remains unreliable for complex 4D tasks whose evidence is distributed across distant moments, scenes, and temporal scales.

Prior work has addressed these limitations through complementary interventions at the system, model, and method levels. At the system level, agentic harnesses iteratively retrieve query-relevant frames, maintain structured temporal and object memories, and orchestrate multi-granular search or specialist tools~\citep{wang2024videoagent, fan2024videoagent, ma2025drvideo, zhang2025deepvideodiscovery}. At the model level, long-video MLLMs extend their effective visual horizon through long-context training, compact per-frame tokenization, and explicit temporal encoding~\citep{chen2025longvila, li2024llamavid, bai2025qwen25vl}. At the method level, adaptive spatiotemporal compression removes redundant frames and visual tokens, sparse memory consolidates dense observations into long-term representations, and query-adaptive hierarchical selection allocates computation across temporal granularities~\citep{shen2025longvu, song2024moviechat, wang2025videotree}. Together, these advances improve access to long-range evidence, but their gains are typically evaluated under different datasets, interfaces, and access conditions, making it difficult to isolate improvements in the base model, evidence acquisition, memory, and final reasoning. Moreover, these components are not readily composable across systems, since methods such as LongVU require access to the internal vision-token pipeline, whereas LongVT learns its tool-use protocol through model-specific post-training of Qwen2.5-VL-7B rather than through a model-agnostic plug-in interface~\citep{shen2025longvu, yang2026longvt}.

Moreover, expanding a model's nominal context window determines how much input it can accept, but not how much distributed evidence it can use effectively, as performance degrades with increasing context length and task complexity even when all relevant evidence is perfectly retrieved~\citep{hsieh2024ruler, du2025contextlength}. External memory and RAG reduce the raw context presented at once, but shift the bottleneck to deciding what to store, compress, update, retrieve, and ultimately use; current systems can lose visual fidelity during memory construction and remain far below human performance on long-term temporal and causal reasoning~\citep{maharana2024locomo, ren2026memlens}. Reusable skills and tools provide procedural guidance and specialist capabilities, but their gains depend strongly on the task and harness, while reliable execution requires correct routing and parameter grounding and deteriorates as tool catalogs, tool outputs, and interaction histories grow~\citep{li2026skillsbench, kate2025longfunceval, shen2026mem2actbench}. Consequently, as task horizons and dependency depth grow by orders of magnitude, errors accumulate across memory construction, retrieval, skill selection, and reasoning, and current systems have yet to demonstrate stable and reliable performance on complex multimodal tasks that demand human-like episodic recall, temporal integration, memory updating, and coordinated action.

\begin{table*}[!t]
  \centering
  \footnotesize
  \setlength{\tabcolsep}{5pt}
  \renewcommand{\arraystretch}{1.15}
  \begin{tabular}{ccccc}
    \toprule[1.5pt]
    Benchmark & Corpus (h) & Temporal extent (Max) & Multi-History Scope? & Harness Support? \\
    \hline
    Video-MME~\citep{fu2025videomme} & 254 & 1 h & \xmark & \xmark \\
    LongVideoBench~\citep{wu2024longvideobench} & NR & 1 h & \xmark & \xmark \\
    HourVideo\citep{chandrasegaran2024hourvideo} & 381 & 2 h & \xmark & \xmark \\
    LVBench~\citep{wang2025lvbench} & 117 & 68.4 min & \xmark & \xmark \\
    InfiniBench~\citep{ataallah2025infinibench} & $>1{,}000$ & 52.6 min & \xmark & \xmark \\
    EgoMonth~\citep{chen2026egomonth} & 301 & ~ 4-5 hours\textsuperscript{a} & \cmark & \cmark \\
    MemLens~\citep{ren2026memlens} & N/A & 32K--256K tokens & \cmark & \cmark \\
    \midrule
    \benchmark{} (Ours) & $\approx\RoundedTotalHours{}$ & \TVMaximumBoundHours{} h & \cmark & \cmark \\
    \bottomrule[1.5pt]
  \end{tabular}
  \vspace{-3.5mm}
  \caption{Comparison with representative long-video and multimodal-memory benchmarks.
  Corpus hours measure total source-video duration, whereas temporal extent describes the video input or history scale, not the minimum evidence needed to answer a question.
  The final column summarizes documented evaluation or harness functionality, rather than whether third-party agents could be integrated.}
  \vspace{-6mm}
  \label{tab:benchmark_comparison}
\end{table*}

Therefore, we introduce LongVideo, a dataset comprising over 1,000 standalone long videos and two multi-season television series, with approximately 900 hours of video in total. Based on LongVideo, we build \dataset{}, which contains more than 5,000 question-answer pairs and combines multi-evidence questions on individual videos with episode-, season-, and complete-series queries across visual, textual, and spatiotemporal reasoning. To support a broad range of LLMs and MLLMs, we further develop the \harness{}. Following the principle of task-independent storage and goal-directed recall, the harness first constructs reusable, question-independent memories of long video streams and then performs question-conditioned retrieval and specialized evidence acquisition when a concrete task arrives. This design decouples the temporal scope of a task from the model's native context window, allowing the foundation model and the harness to operate together as the \method{}.

Our contributions are summarized as below:

\noindent 1. We curate the \dataset{} and the \benchmark{}, which is, to our knowledge, the first multimodal video benchmark in which an individual question can bind more than 100 hours of ordered source video. The benchmark contains 1,000+ videos, 5,000+ manually reviewed QA pairs, and 850+ hours of source video across visual, textual, and spatiotemporal reasoning.
  
\noindent 2. We develop the \method{}, a model-flexible inference framework that integrates multiple complementary skills and tools, a memory-centric RAG system, and more than 20 supporting models. It constructs reusable question-independent memories and performs question-conditioned retrieval and specialized evidence acquisition, decoupling task duration from the native context window of the underlying LLM or MLLM.
  
\noindent 3. Through a comprehensive evaluation of leading LLMs and MLLMs, we show that standalone models struggle with complex long-horizon 4D tasks, while harness-equipped \method{} consistently improve their performance across task types and temporal scales.

\begin{figure}[!t]
    \centering
    \vspace{-15mm}
    \includegraphics[width=\linewidth]{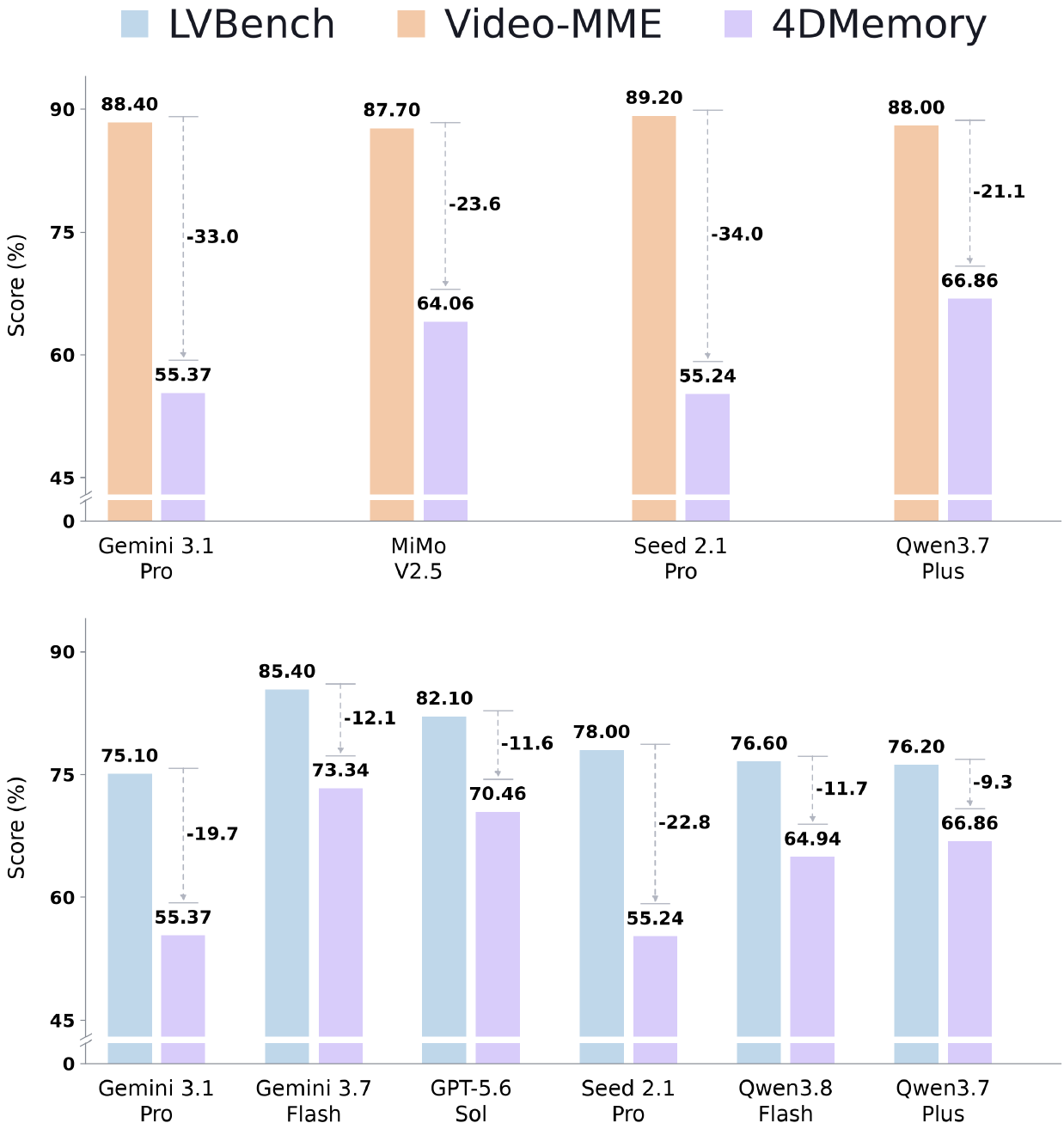}
    \vspace{-20mm}
    \caption{Comparison of previous Video-MME (top) and LVBench (bottom) scores with 4DMemory results for the same models. All models shown score lower on 4DMemory, with dashed arrows indicating the decrease in percentage points.}
    \vspace{-6mm}
    \label{fig:comparison}
\end{figure}

\vspace{-2.5mm}
\section{The 4DMemory Benchmark}
\label{sec:benchmark}
\vspace{-2.5mm}

\begin{figure*}[!t]
  \centering
  \includegraphics[width=\linewidth]{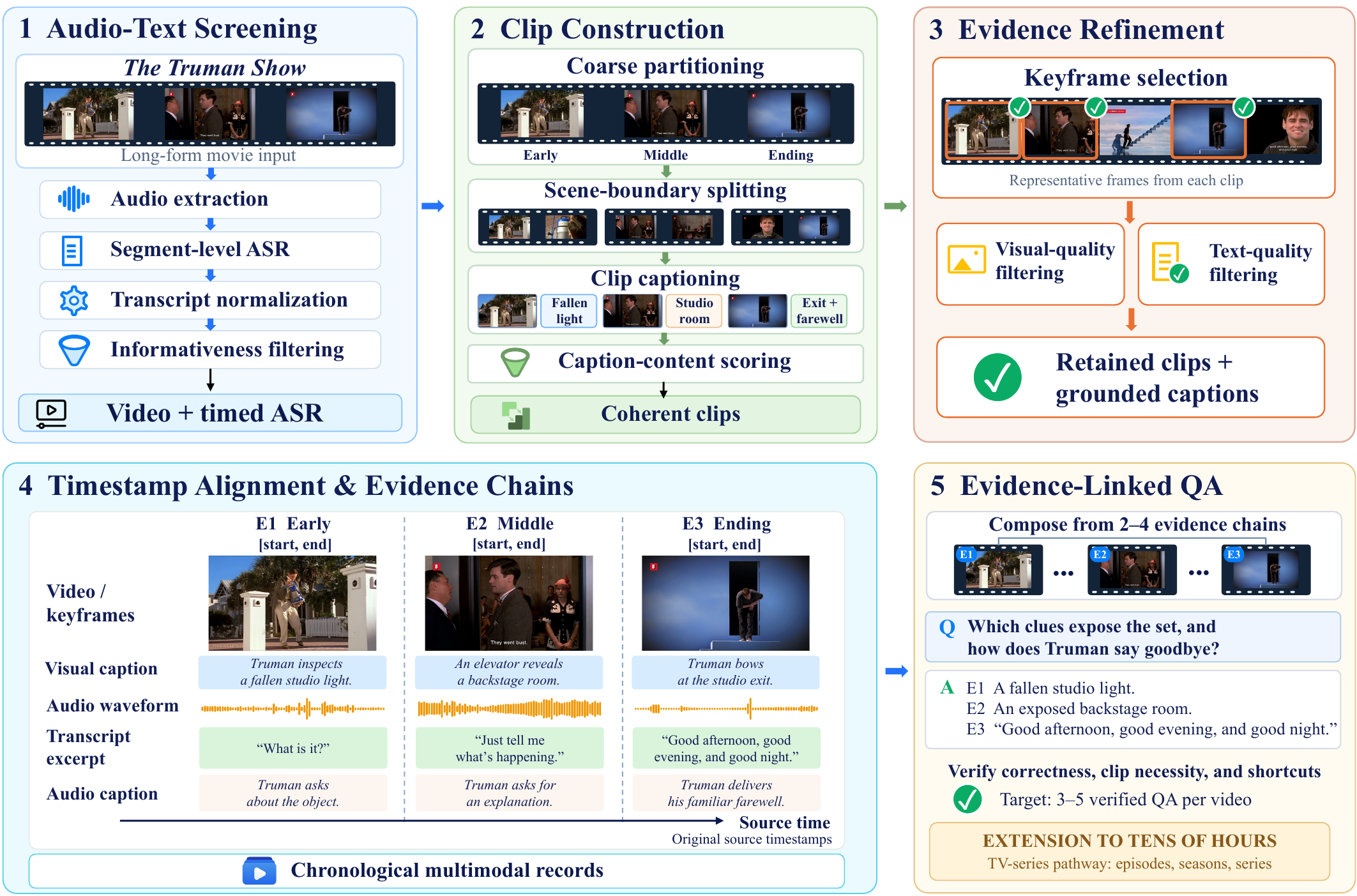}
  \vspace{-8mm}
  \caption{Overview of the LongVideo construction pipeline. Hours-level videos undergo audio-text screening, clip construction, and evidence refinement to obtain coherent clips with grounded captions. Visual evidence, audio captions, and transcripts are aligned to original source timestamps and organized into chronological multimodal evidence chains. QA pairs are then constructed from multiple evidence chains and checked for answer correctness, clip necessity, and potential shortcuts.}
  \vspace{-5mm}
  \label{fig:dataset}
\end{figure*}

Existing long-video benchmarks primarily evaluate reasoning over individual videos lasting one to two hours. Even for LLMs and MLLMs with context windows exceeding 1M tokens, these settings alone do not establish whether models can match human abilities to extract and retain fine-grained evidence, or preserve, integrate, and use key information after receiving tens of hours of additional video. Recent work has begun to evaluate cross-day recall from egocentric video~\citep{chen2026egomonth}, but its source content remains relatively homogeneous and its task coverage limited. We therefore construct \dataset{} and \benchmark{} using temporally connected videos and multi-season television series to evaluate visual, linguistic, and spatiotemporal reasoning grounded in long-horizon memory. As shown in \Cref{tab:benchmark_comparison}, \dataset{} extends prior benchmarks along four dimensions: corpus duration, question-level temporal coverage, the scope of multi-evidence integration, and harness support\footnote{\scriptsize NR: no corpus-hour total stated in the cited source; N/A: not a video corpus. \textsuperscript{a}reports calendar recording spans per participant, not continuous video hours or maximum per-question input duration. MemLens evaluates image--text conversations, and its memory-agent comparison uses a 195-question subset.}.

\paragraph{Task definition.}
In \benchmark{}, each item can be represented as
\begin{equation}
  x_i = (d_i, q_i, \tau_i, \mathcal{V}_i; a_i, \mathcal{E}_i, p_i, c_i),
  \label{eq:item_schema}
\end{equation}
where $d_i$ is a unique identifier, $q_i$ is the question, $\tau_i$ specifies the answer format, and $\mathcal{V}_i=(v_{i,1},\ldots,v_{i,n_i})$ is the ordered set of source videos. The private annotations comprise the reference answer $a_i$, supporting evidence $\mathcal{E}_i$, an optional executable program $p_i$, and task metadata $c_i$. Given $q_i$, $\tau_i$, and access to the complete scope $\mathcal{V}_i$, a model must produce $\hat{a}_i$ without these private annotations. Standalone questions connect distant moments within one video ($n_i=1$), whereas serialized questions bind the ordered episodes of an episode-, season-, or series-level scope. Each item is assigned to one mutually exclusive task family: \emph{visual} perception and aggregation, \emph{text} understanding grounded in linguistic evidence, or \emph{spatiotemporal} reasoning about relations, order, state changes, and cross-modal dependencies. The task label is used for reporting, not provided as an oracle tool route. Here, 4D reasoning concerns evolving visual and image-plane spatial evidence, rather than metric 3D reconstruction.

\vspace{-2.5mm}
\subsection{Dataset collection and processing}
\label{sec:dataset_collection}
\vspace{-2.5mm}

\paragraph{Source collection.} To broaden the corpus and evaluate current models across diverse knowledge domains and tasks, we construct our dataset from two complementary sources with different temporal structures. Our dataset comprises \StandaloneVideoCount{} standalone web videos and \TVEpisodeCount{} television episodes from \TVSeriesCount{} series, totaling \TotalHours{} hours of distinct source video. Standalone videos support reasoning along continuous timelines, while serialized episodes extend memory across recurring characters and settings, with the longest complete-series scope reaching \TVMaximumBoundHours{} hours. Detailed distributions of video content categories and QA task types are presented in Appdendix \cref{fig:dataset_distribution}.

\begin{figure*}[!t]
  \centering
  \includegraphics[width=\linewidth]{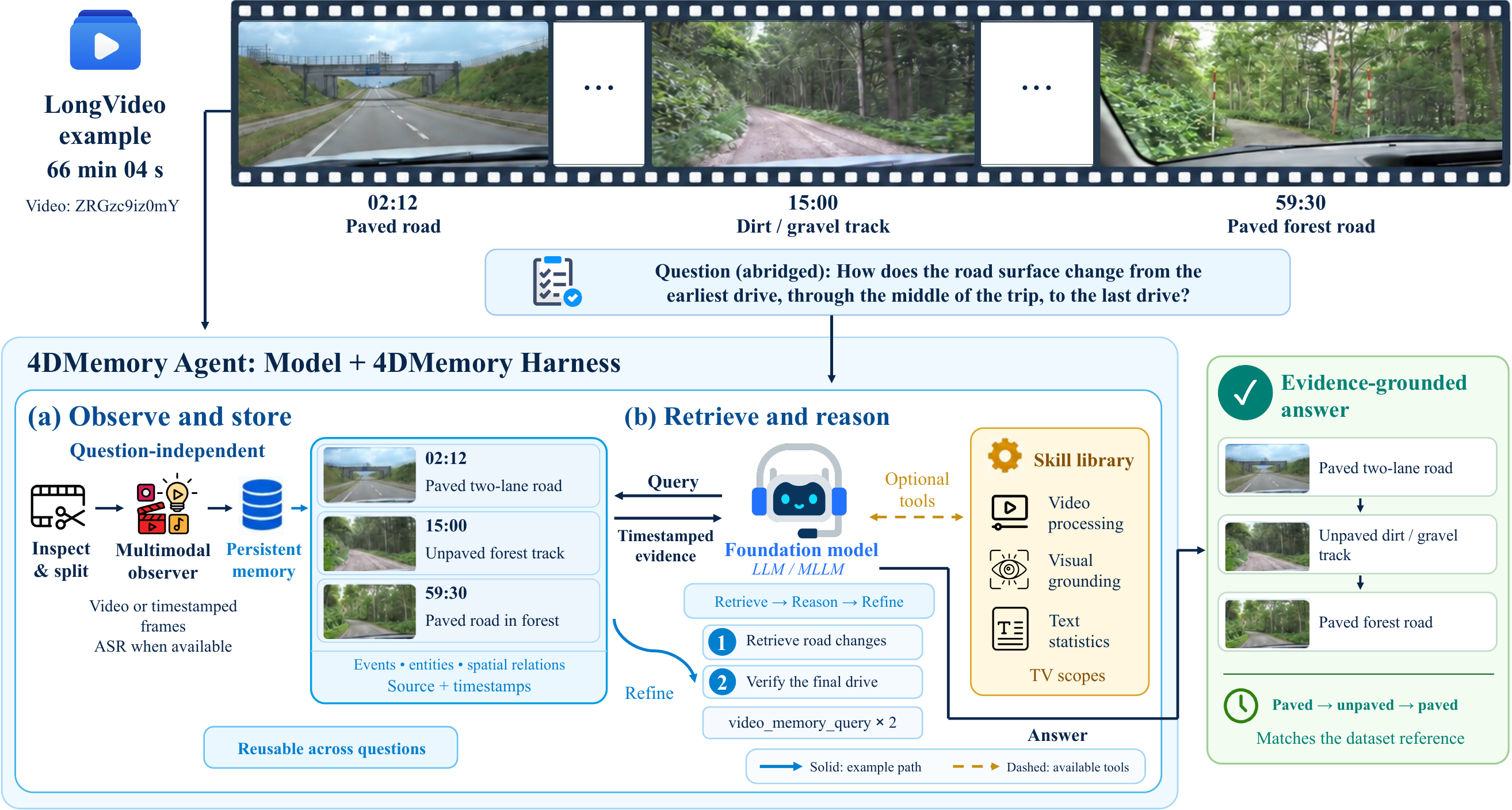}
  \vspace{-8mm}
  \caption{Illustration of 4DMemory Agent framework. (a) The harness builds persistent, question-independent memory of events, entities, and spatial relations, preserving source timestamps for reuse across questions. (b) A foundation model retrieves timestamped evidence and iteratively refines its queries and reasoning, with optional tools provided by a skill library.}
  \vspace{-5mm}
  \label{fig:method}
\end{figure*}

\vspace{-2.5mm}
\paragraph{Aligned multimodal records.}
As shown in \cref{fig:method}, we organize video, audio, and textual annotations into chronological records linked by source identifiers and timestamps. For standalone videos, each record associates sampled frames and visual captions with the corresponding audio segments and available ASR transcripts. For television sources, multimodal annotation additionally provides scene summaries, speaker-labelled dialogue, and utterance-level temporal boundaries. Segment-local timestamps are mapped back to the original source timeline to preserve temporal correspondence across modalities. This representation supports both continuous video timelines and episode--season--series hierarchies. Records passing a unified quality threshold are retained for compositional QA construction and verification, while reference answers, annotated evidence locations, and reasoning programs remain private during evaluation.
Preprocessing and alignment details can be referred in the \cref{subsec:data_preprocess}.

\vspace{-2.5mm}
\subsection{Compositional QA construction}
\label{sec:qa_construction}
\vspace{-2mm}
We turn the aligned records into questions that require combining evidence, rather than recalling an isolated fact. For standalone videos, the generator receives time-ordered evidence cards containing clip summaries, available ASR, and temporal bounds. Each candidate connects \EvidenceClipMinimum{} to \EvidenceClipMaximum{} clips through an explicit operation such as comparison, temporal ordering, state change, cross-modal binding, or distributed counting. A separate verifier checks the answer and the joint necessity of the evidence by removing each supporting clip in turn. Only \RetainedQAMinimum{} to \RetainedQAMaximum{} accepted questions are retained per video, yielding \StandaloneQACount{} QA pairs.

For serialized video, we first compute reference answers with deterministic programs over validated dialogue, ASR, and visual statistics at the requested scope. A language model rewrites each program as a self-contained question, and an independent verification call checks that the wording preserves the program's meaning and answer. This produces \TVQACount{} questions covering episode, season, and complete-series aggregation, comparison, ordering, and trend analysis. Visual counts denote tracker identities within a clip, summed across clips, rather than globally deduplicated people or objects.

Both pathways apply clarity, evidence-sufficiency, compositionality, and difficulty checks; standalone candidates additionally undergo question-only shortcut screening. The current release excludes \TVPrunedQACount{} TV questions affected by invalid zero-duration visual clips and contains \TotalQACount{} globally unique QA pairs. A shared schema assigns every question to exactly one task partition, and release validation checks identifiers, partition consistency, file sizes, and content hashes.
Construction details are provided in \cref{subsec:qa_construct}.

\vspace{-2.5mm}
\subsection{Compositional QA construction}
\label{sec:standalone_construction}
\vspace{-2mm}
To evaluate reasoning over evidence distributed across time, we construct compositional questions that require an explicit relation or operation. For standalone videos, the generator receives time-ordered evidence cards containing clip summaries, available ASR transcripts, and temporal bounds. Candidates combine \EvidenceClipMinimum{} to \EvidenceClipMaximum{} clips through operations such as comparison, temporal ordering, state change, cross-modal binding, or distributed counting. A separate verifier checks answer correctness and evidence sufficiency, and assesses each clip's necessity by judging whether the complete answer would remain derivable without it. Each retained video contributes \RetainedQAMinimum{} to \RetainedQAMaximum{} accepted questions, yielding \StandaloneQACount{} QA pairs.

For television sources, deterministic programs derive reference answers from dialogue, ASR, and clip-level visual statistics within the requested temporal scope. Questions are instantiated from fixed templates or written by a language model to express the specified computation. Model-authored questions undergo separate verification of their consistency with the underlying program, while the long-duration extension additionally uses blind expression reconstruction and program-equivalence checks. The resulting \TVQACount{} questions cover aggregation, comparison, ordering, and trend analysis across episode, season, and series scopes. Visual counts represent tracked instances within individual clips and are summed across clips, so the same person or object may contribute multiple times across the source timeline.

We combine deterministic structural and program checks with model-based assessments of clarity, evidence support, compositionality, and difficulty. Question-only shortcut audits additionally screen standalone candidates and original TV rewrites for answers recoverable from the question wording or prior knowledge. After excluding \TVPrunedQACount{} TV questions affected by invalid zero-duration visual clips, the current release contains \TotalQACount{} globally unique QA pairs. Each question is assigned to exactly one task partition under a shared schema, and release validation checks identifier uniqueness, partition consistency, file sizes, and content hashes. Further construction details are provided in \cref{subsec:qa_construct}.

\vspace{-2.5mm}
\subsection{4DMemory Harness}
\label{sec:harness}
\vspace{-2mm}

Complex tasks over long multimodal histories require both access to diverse evidence and selective recall of earlier observations. Although modern LLMs can compress context and invoke skills, these capabilities do not remove the constraints imposed by native input support and finite context windows. Repeated compression may also omit details that become relevant to subsequent questions. We therefore introduce \harness{}, which separates media processing, persistent memory, and question-conditioned reasoning. The harness exposes three processing skills---video processing, visual grounding, and text statistics---and stores question-independent, timestamped observations in persistent structured memory. A lightweight retrieval layer selects relevant records and their temporal neighbors to provide bounded evidence for reasoning. When the visual input pathway does not carry audio, an internal transcription module processes the available source audio and incorporates timestamped speech evidence into the corresponding memory records.

% \yh{Refined Here.}
\vspace{-2.5mm}
\paragraph{Video-processing skill.}
The video-processing skill prepares bound sources for model observation through size inspection and temporal segmentation under request constraints. It preserves source-time references so that evidence from separate chunks can be related along the original timeline.

\vspace{-2.5mm}
\paragraph{Visual-grounding skill.}
The visual-grounding skill uses SAM 3.1 to provide tracking evidence for supported TV scopes. It returns ordered person and object counts within episodes and person-count aggregates over longer scopes, supporting counting, comparison, and temporal reasoning.

\vspace{-2.5mm}
\paragraph{Text-statistics skill.}
The text-statistics skill provides deterministic counts, speaker rankings, and aggregates derived from TV dialogue and ASR annotations. A complementary episode-analysis tool produces bounded semantic summaries and topic information when richer textual evidence is needed. 

All three skills expose typed operations, with source configuration and execution policies controlled by \harness{}.

\begin{table*}[!t]
    \centering
    \footnotesize
    \setlength{\tabcolsep}{3pt}
    \renewcommand{\arraystretch}{1.25}
    \begin{tabular}{@{}l*{7}{r}@{}}
        \toprule[1.5pt]
        & \shortstack{Invocation}
        & \multicolumn{2}{c}{\shortstack{Ordered}}
        & \multicolumn{2}{c}{\shortstack{Order-relaxed}}
        & \multicolumn{2}{c}{\shortstack{Overall}} \\
        \cmidrule(lr){3-4}\cmidrule(lr){5-6}\cmidrule(l){7-8}
        Model & & 3 judges & Opus-5 & 3 judges & Opus-5 & 3 judges & Opus-5 \\
        \midrule
        Gemini 3.1-Pro & 99.81 & 58.38 & 63.82 & 58.38 & 63.82 & 58.27 & 63.69 \\
        GPT-5.6 Sol & 100.00 & 70.14 & 71.29 & 70.46 & 71.61 & 70.46 & 71.61 \\
        Opus-4.8 & 98.70 & 71.96 & 73.22 & 72.76 & 74.02 & 71.81 & 73.05 \\
        MiMo-V2.5 & 100.00 & 58.59 & 60.84 & 64.06 & 66.36 & 64.06 & 66.36 \\
        Grok-4.6 & 100.00 & 72.04 & \underline{74.11} & \underline{75.55} & \underline{77.64} & \underline{75.55} & \underline{77.64} \\
        Muse Spark-1.1 & 99.88 & 70.89 & 71.79 & \textbf{77.88} & \textbf{78.82} & \textbf{77.79} & \textbf{78.73} \\
        Seed 2.1-Pro & 99.69 & 44.61 & 45.48 & 55.41 & 56.29 & 55.24 & 56.12 \\
        Gemini 3.7-Flash & 100.00 & \textbf{73.34} & \textbf{74.66} & 73.34 & 74.66 & 73.34 & 74.66 \\
        GLM 5.3-Flash & 99.81 & 55.76 & 57.15 & 65.20 & 66.60 & 65.07 & 66.47 \\
        Qwen3.8-Max & 99.87 & \underline{72.69} & 73.67 & 72.69 & 73.67 & 72.59 & 73.57 \\
        DeepSeek V4-Pro & 99.85 & 68.69 & 69.36 & 68.69 & 69.36 & 68.58 & 69.26 \\
        Qwen3.8-Flash & 94.13 & 63.00 & 63.55 & 68.99 & 69.58 & 64.94 & 65.50 \\
        Qwen3.7-Plus & 99.81 & 57.72 & 60.34 & 66.99 & 69.64 & 66.86 & 69.51 \\
    \bottomrule[1.5pt]
    \end{tabular}
    \vspace{-3mm}
    \caption{The overall performance on the 4DMemory Benchmark among 13 SOTA methods.}
    \vspace{-6mm}
    \label{tab:main_answer_results}
\end{table*}

\vspace{-2.5mm}
\paragraph{Multimodal observation.}
After media preparation, an observer converts temporal units into structured observations without access to the benchmark question. It receives native video or timestamped frames according to the configured input pathway. When enabled, an internal transcription module adds timestamped speech evidence from the source audio.

\vspace{-2.5mm}
\paragraph{Question-independent memory construction.}
Observations are stored as persistent documents with source identities, temporal bounds, and structured multimodal fields. Their construction excludes question-specific supervision, allowing provenance-matched records to be reused across questions and temporal scopes. Memory reuse follows the configuration of the actual observer, including any separately configured observer used by a text-only reasoner.

\vspace{-2.5mm}
\paragraph{Retrieval and memory-guided reasoning.}
At question time, \harness{} retrieves bounded evidence from the documents associated with the current item. The reasoner can request additional evidence through \texttt{video\_memory\_query} and invoke available specialists to support its answer. Final reasoning operates on the question, retrieved memory, and tool outputs, separating the stored history from the model's active context. Observation, retrieval, and execution details are provided in \cref{subsec:harness}.

\vspace{-2.5mm}
\section{Experiments}
\label{sec:experiments}
\vspace{-2.5mm}

\paragraph{Experimental setup.}
We evaluate 13 API-served models on all 5,214 questions in \benchmark{} using \harness{}, without benchmark-specific parameter updates. Each system uses Gemini 3.1 Pro as a separate media observer, enabling evaluated models to access evidence from modalities they cannot process directly. To reduce dependence on a single judge, we assess the same model responses under two independent judging protocols: an Opus-5 judge and a majority-vote panel comprising Kimi K3, GLM 5.3, and Qwen3.8-Max. Tool invocation and execution are assessed separately, and performance is summarized through four complementary metrics. \emph{Invocation} measures the fraction of questions with at least one accepted tool call, while \emph{Ordered} measures the fraction of these questions answered correctly with all required operations successfully executed in the prescribed order. \emph{Order-relaxed} removes the execution-order constraint, while \emph{Overall} measures the resulting joint success rate over all benchmark questions.

% \Cref{tab:main_answer_results} reports aggregate performance on \EvaluationSnapshotQACount{} questions after supplementary judging of 205 historical Gemini-3.1 answers, without rerunning any evaluated model.
% The six task-by-scope and subtype tables, their scoring definitions, and the corresponding analyses in \cref{app:full_results} retain the \EvaluationSnapshotTime{} snapshot of current primary answers from \EvaluationSnapshotDate{}, without historical best-of selection.
% \Cref{sec:result_snapshot} explains the shared question counts and their relation to the earlier \TotalQACount{}-question release.

\begin{table*}[!t]
  \centering
  \footnotesize
  \setlength{\tabcolsep}{3pt}
  \renewcommand{\arraystretch}{1.25}
  \begin{tabular}{@{}lcccccc@{}}
    \toprule[1.5pt]
    Model & Ordering & State & Counting & Trajectory & CM reason & CM bind \\
    \hline
    Gemini 3.1-Pro & 68.85 & 50.18 & 30.91 & 42.86 & 55.43 & 61.18 \\
    GPT-5.6 Sol & 72.49 & 67.16 & 44.35 & 57.14 & 70.11 & 72.37 \\
    Opus-4.8 & 75.40 & 66.42 & \underline{45.97} & 50.93 & \underline{75.54} & 76.97 \\
    MiMo-V2.5 & 71.76 & 55.90 & 36.02 & 43.48 & 65.22 & 57.24 \\
    Grok-4.6 & \textbf{83.19} & \underline{69.74} & 45.70 & 54.66 & \textbf{78.80} & \textbf{81.58} \\
    Muse Spark-1.1 & \underline{82.02} & \textbf{73.80} & \textbf{47.58} & \underline{62.73} & \textbf{78.80} & \underline{78.95} \\
    Seed 2.1-Pro & 67.25 & 51.29 & 28.23 & 38.51 & 55.98 & 56.58 \\
    Gemini 3.7-Flash & 82.17 & 66.61 & 45.43 & 53.42 & 72.83 & 78.29 \\
    GLM 5.3-Flash & 76.06 & 55.54 & 34.41 & 47.20 & 63.59 & 69.74 \\
    Qwen3.8-Max & 78.89 & 69.37 & 40.32 & \textbf{66.46} & 69.02 & 71.71 \\
    DeepSeek V4-Pro & 77.58 & 60.15 & 31.72 & 49.69 & 64.67 & 68.42 \\
    Qwen3.8-Flash & 69.87 & 68.08 & 37.90 & 57.76 & 67.93 & 71.71 \\
    Qwen3.7-Plus & 73.00 & 54.24 & 39.25 & 42.86 & 67.93 & 67.11 \\
    \bottomrule[1.5pt]
  \end{tabular}
  \vspace{-3.5mm}
  \caption{Six selected standalone-video spatiotemporal subtypes. Columns denote multi-event ordering, state-change tracking, distributed counting, path and trajectory tracking, cross-modal reasoning, and cross-modal binding, respectively.}
  \vspace{-6.5mm}
  \label{tab:video_temporal_subtypes}
\end{table*}

\vspace{-2.5mm}
\paragraph{Skills/Tools is crucial for LLMs in complex tasks.} The overall results in \cref{tab:main_answer_results} show that frequent tool invocation leaves substantial room for improving joint success. Twelve of the 13 systems invoke tools on at least 98.70\% of questions, yet the highest observed Overall score is 77.79\%, achieved by Muse Spark-1.1, followed by Grok-4.6 (75.55\%) and Gemini 3.7-Flash (73.34\%). Execution-order requirements also change the ranking: Gemini 3.7-Flash leads Ordered at 73.34\%, whereas Muse scores 70.89\% Ordered and 77.88\% Order-relaxed, both conditional on invocation. The latter difference corresponds to 364 correctly answered questions with successful required operations but an ordering violation. These results distinguish initiating tool use, satisfying execution requirements, and reaching a correct answer, motivating evaluation of both the process and its outcome.

\vspace{-2.5mm}
\paragraph{Spatiotemporal task needs fine-grained reasoning ability.} The standalone-video subtype results reveal a consistent weakness in distributed counting across six subtypes covering 2,785 questions. Every system has a lower observed counting score than ordering score, and the best counting result is only 47.58\% (Muse; 177/372), with complete judging. This gap also holds for fully evaluated systems: GPT-5.6 Sol scores 72.49\% on ordering versus 44.35\% on counting, while Gemini 3.7-Flash scores 82.17\% versus 45.43\%. Thus, ordering events does not ensure reliable accumulation of counts across separated observations. Other subtypes expose complementary strengths: Grok leads observed ordering and cross-modal binding at 83.19\% and 81.58\%, Muse leads state-change tracking at 73.80\%, and Qwen3.8-Max leads trajectory tracking at 66.46\%. Some leading entries remain provisional, as marked in \ref{tab:video_temporal_subtypes}. Together, these results identify both a shared counting weakness and distinct capability profiles that aggregate accuracy would obscure. 

Aggregate margins reflect task composition. Standalone spatiotemporal questions constitute 77.46\% of the benchmark, yet this group does not explain Muse’s lead over Grok: their scores are nearly identical at 74.92\% and 74.97\%. Across all standalone videos, Grok leads by eight successes, whereas Muse leads by 125 on television questions, giving a net advantage of 117. Episode-level visual questions alone contribute a difference of 116 successes, with Muse solving 163/172 and Grok solving 47/172. The aggregate lead therefore should not be interpreted as a uniform advantage across task groups.

Temporal scope does not produce a uniform performance trend. GPT-5.6 Sol’s television spatiotemporal accuracy decreases from 87.86\% at episode scope to 38.98\% at season scope and 12.50\% at complete-series scope. Gemini 3.7-Flash shows the opposite pattern: 72.86\%, 84.75\%, and 100.00\%. Both systems have complete judging in these groups. Since the groups differ in reasoning operations and evidence requirements as well as temporal extent, the results do not isolate an effect of increasing duration. Temporal scope must therefore be interpreted together with task composition.

Local comparisons and sequence-wide aggregation separate sharply. GPT solves all 64 person-minus-object comparisons and 107/108 person-to-object ratios, but only 1/17 above-mean unit-count questions. Similarly, it solves all 64 first-appearance ordering and localization questions, but only 1/16 temporal-half comparisons and 1/17 trend-reversal questions. Grok exhibits a different profile: 25.93\% on person-to-object ratios, compared with 82.35\% on above-mean counting and 94.12\% on trend reversals. These differences distinguish retrieving selected occurrences and comparing quantities from aggregating an ordered sequence. Joint accuracy alone cannot isolate retrieval, execution, or computation as the cause of failure. Additional task-by-scope analyses and evaluation qualifications are provided in \cref{app:full_results}.

\vspace{-1.5mm}
\section{Conclusion}
\label{sec:conclusion}
\vspace{-1mm}
We introduced \dataset{} and \benchmark{} to evaluate long-horizon multimodal memory across dozens of hours of video. The benchmark comprises approximately \RoundedTotalHours{} hours of video and covers visual, textual, and spatiotemporal reasoning. Our \harness{} combines question-independent persistent memory, bounded retrieval, and specialized tools, with separate metrics for tool invocation, execution order, and joint answer success. Across 13 evaluated models, high invocation rates coexist with substantial differences in workflow compliance and final-answer correctness, with the highest observed joint success reaching 77.79\% under judging. Task-level analyses reveal persistent difficulties in distributed counting and multi-scope aggregation, while differences across sources and temporal scopes expose complementary system strengths. These findings establish a basis for studying how observation, memory access, and tool-mediated computation jointly support reliable reasoning over extended multimodal histories.

\clearpage

\section*{Limitations}
\label{app:limitations}
Our study does not include agent fine-tuning on our data, leaving the potential benefits of such adaptation unexplored. Consequently, the reported results do not establish the effectiveness of these data as a training resource for agents. Evaluation across additional downstream tasks is also beyond the scope of this work, and the broader applicability of our approach remains to be established. Tasks with different objectives and interaction settings may require further adaptation and dedicated evaluation.
\bibliography{references}

\clearpage
% \onecolumn
\appendix

\setcounter{figure}{0}  % 图像从 0 计数，下一张为 Figure 1
\setcounter{table}{0}  % 图像从 0 计数，下一张为 Figure 1
\renewcommand{\thesection}{S\arabic{section}}
\renewcommand{\thesubsection}{S\arabic{section}.\arabic{subsection}}
\renewcommand{\thetable}{S\arabic{table}}
\renewcommand{\thefigure}{S\arabic{figure}}

\noindent {\LARGE \textbf{Supplementary Material}}

\section{Related Work}
\label{sec:related_work}
% \vspace{-4mm}
\paragraph{Long-video benchmarks.}
Long-video benchmarks increasingly test evidence integration over extended temporal spans~\citep{zhu2026chronophybench}. EgoSchema measures intrinsic evidence span with temporal certificate sets, while MMBench-Video uses free-form questions over multi-shot videos~\citep{mangalam2023egoschema, fang2024mmbenchvideo}. LongVideoBench and Video-MME extend evaluation to hour-long inputs and multiple modality streams, while MLVU broadens video genres and task formats~\citep{wu2024longvideobench, fu2025videomme, zhou2025mlvu}. TempCompass and VSI-Bench isolate fine-grained temporal reasoning and spatial memory, respectively~\citep{liu2024tempcompass, yang2025vsibench}. At longer scales, LVBench evaluates multi-hour videos and InfiniBench covers long-form movies and television episodes, highlighting persistent difficulty in temporal grounding and cross-event reasoning~\citep{wang2025lvbench, ataallah2025infinibench, wen2026vulnerabilities}. EgoMonth extends this direction to month-level per-participant continuity and cross-video egocentric QA over a 301-hour longitudinal corpus~\citep{chen2026egomonth}.
% \vspace{-4mm}
\paragraph{Agentic long-video understanding.}
Agentic approaches treat long-video understanding as iterative evidence acquisition through observation selection, tool use, and reasoning-state updates. The two VideoAgent systems exemplify query-driven frame retrieval and structured temporal or object memory, respectively~\citep{wang2024videoagent, ataallah2025infinibench, jia2026seeing, jia2026mmoe}. Retrieval- and planning-centered systems convert videos into searchable evidence, zoom across space and time, and control the scope and density of each observation~\citep{ma2025drvideo, ye2025temporalsearch, zhang2025deepvideodiscovery, li2026lenswalk}. Memory-centric agents organize episodic, semantic, visual, or hierarchical representations for iterative retrieval and reasoning~\citep{yeo2026worldmm, yin2026videoarm}. LongVT and VideoThinker internalize multi-step tool use through model-specific post-training or synthetic tool-interaction trajectories~\citep{yang2026longvt, li2026videothinker}. Beyond video-specific systems, Skill Retrieval Augmentation decomposes scalable skill use into retrieval, incorporation, and end-task execution, while AgenticSTS uses typed retrieval to bound long-horizon decision contexts and isolate memory layers~\citep{li2026skillsbench, cheng2026agenticsts}. More generally, EnvHarness reshapes static agent environments through Stage, Contract, and Chain plug-ins while retaining their original verifiers, and EnvRigger uses black-box policy rollouts to diagnose weaknesses and validate targeted customizations~\citep{huang2026envharness}. MemoHarness instead adapts six harness control dimensions from accumulated execution experience and reports selective transfer across unseen task suites and base models~\citep{huang2026memoharness}.
% \vspace{-4mm}

\paragraph{Multimodal long-term memory.}
Long-term memory research mainly evaluates sustained, multi-session interactions~\cite{zhao2026unifle}. LoCoMo covers question answering, event summarization, and multimodal dialogue generation, while LongMemEval tests extraction, cross-session and temporal reasoning, knowledge updates, and abstention~\citep{maharana2024locomo, wu2025longmemeval}. MMRC evaluates memory abilities in real-world multimodal conversations, whereas MemLens compares long-context vision-language models with memory-augmented agents~\citep{xue2025mmrc,ren2026memlens}. Mem-Gallery assesses memory extraction, reasoning, and knowledge management, while M3-Agent builds entity-centric episodic and semantic memories from continuous visual and auditory observations~\citep{bei2026memgallery, long2026m3agent, jia2025uni, wu2025query, jiatowards}. Mem2ActBench extends evaluation beyond recall to using stored information for tool selection and parameter grounding~\citep{shen2026mem2actbench}. MemoryArena couples memory formation with action across interdependent sessions and exposes failures missed by recall-oriented benchmarks~\citep{he2026memoryarena}. Harness the Memory compares heterogeneous memory substrates under a common evaluation harness and finds that their effectiveness depends on task regime, retrieval breadth, and history scale~\citep{huang2026envharness}.

\section{Dataset \& Harness Details}
The standalone collection comprises \StandaloneVideoCount{} web videos sourced from Qwen-Planner-Agent's datasets~\cite{qu2026qwen}, including \BaseVideoCount{} base videos and \ExtensionVideoCount{} extension videos. Sources are deduplicated by identifier and selected for compositional QA construction following \cref{sec:qa_construction}, with durations ranging from \StandaloneMinDurationMinutes{} minutes to \StandaloneMaxDurationHours{} hours. Content covers crafts and DIY, food and cooking, travel and culture, nature and environment, games and hobbies, science and technology, programming and software, business and finance, and history, politics, and law. The serialized collection comprises \TVEpisodeCount{} episodes from the Chinese-language series \emph{iPartment} and the English-language series \emph{The Big Bang Theory}, spanning \TVSeasonCount{} seasons and \TVHours{} hours. Their story themes include romantic relationships, marriage and family life, friendships, and workplace experiences. We preserve episode identifiers, original timestamps, and chronological order. For composition analysis, content labels are derived from standalone video descriptions and episode-level story summaries, with semantically related low-frequency categories merged for reporting.

\subsection{QA construction}
\label{subsec:qa_construct}
The resulting benchmark contains 5,214 question--answer pairs, comprising 826 TV questions and 4,388 Video questions, designed to assess evidence retrieval, aggregation, and relational reasoning across temporal scales. TV questions span individual episodes, seasons, and entire series, with a single question covering up to \TVMaximumBoundEpisodes{} episodes. Their 42 task subtypes fall into three families: visual perception and aggregation (27.4\%), textual understanding and statistics (45.6\%), and spatiotemporal reasoning (27.0\%). Representative tasks include person--object count comparisons, speaker-level dialogue statistics, first-appearance ordering, and tracking temporal changes across episodes. Video questions cover 12 task subtypes that require integrating evidence from multiple clips within a continuous timeline, including multi-event ordering, multi-evidence comparison, state-change tracking, distributed counting, and cross-modal reasoning.

\subsection{Multimodal preprocessing and alignment.}
\label{subsec:data_preprocess}
For standalone videos, scene-change detection determines the initial clip boundaries, with additional splitting for scenes exceeding the configured duration limit. Audio is extracted and transcribed only for clips containing audio tracks, while sampled frames are used to generate visual captions. For television episodes, coarse segments are formed by placing boundaries at ASR sentence endings closest to the target partition times. Multimodal analysis subsequently divides these segments into coherent units according to changes in scene, dialogue topic, and speaker interaction. Where available, the audio-analysis branch additionally produces semantic summaries, topic and emotion annotations, and audio-quality assessments.

Post-processing merges consecutive utterances attributed to the same speaker, retaining the first utterance's start time and the last utterance's end time. Both clip and utterance timestamps are converted from segment-local coordinates to the original source timeline by adding the parent segment's start-time offset. The resulting intervals guide the extraction of fine-grained video clips and utterance-level audio segments, together with frame sampling from the video clips. For television sources, this conversion is performed within each episode, preserving its original temporal reference. An optional calibration utility matches annotated utterances to ASR segments using text similarity and can further refine their boundaries using voice activity detection. The selected calibration method is recorded in the processing metadata.

Quality screening uses a shared rubric covering description accuracy, temporal coherence, detail richness, and language quality. Each candidate record receives an overall quality score, and a single predefined acceptance threshold is applied across standalone and television sources. Records below this threshold are excluded before QA construction.

\begin{figure*}[!t]
    \centering
        \includegraphics[width=\linewidth]{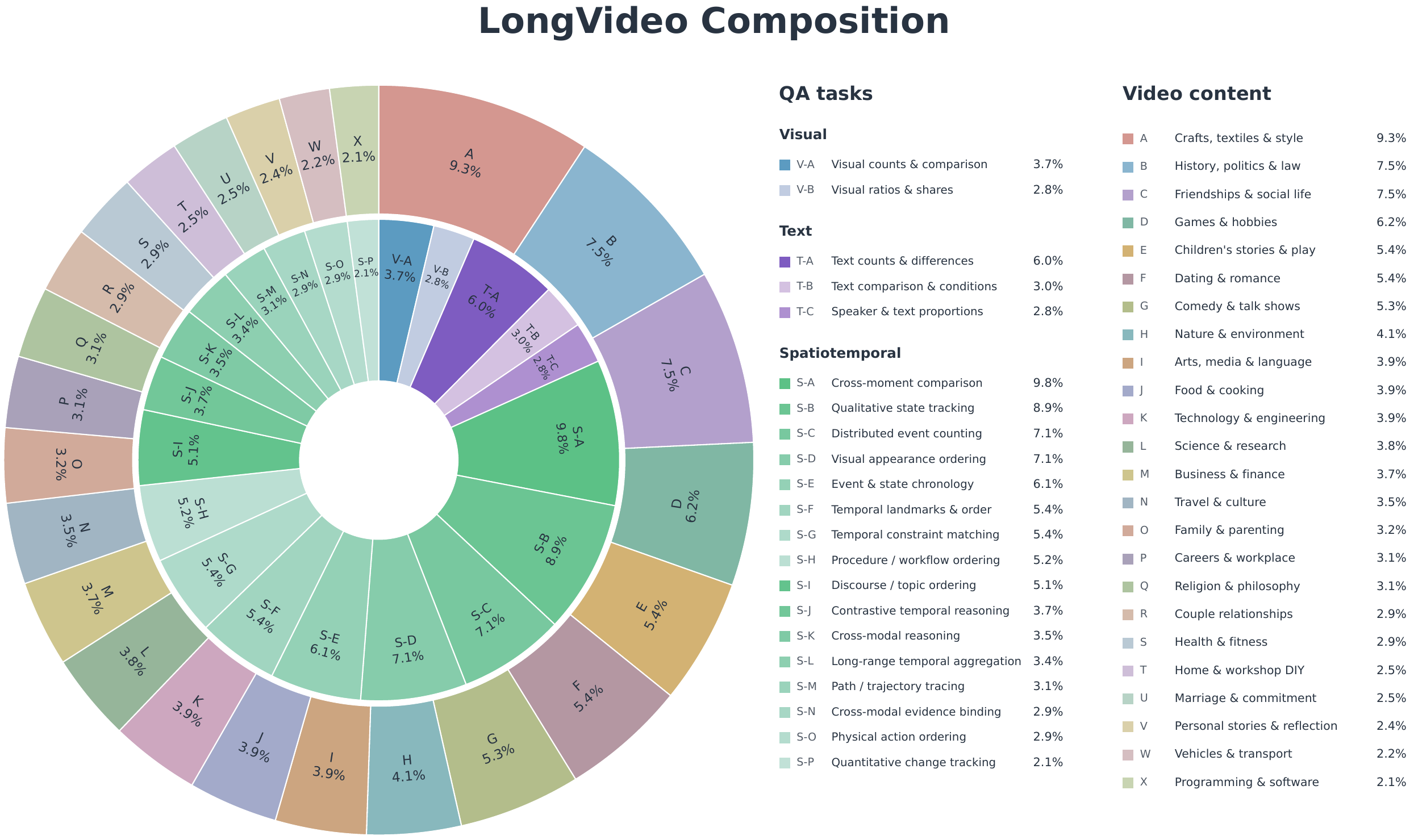}
    \caption{
        The composition details of LongVideo dataset. The inner ring shows the distribution of QA pairs across fine-grained tasks, with blue, purple, and green representing visual, textual, and spatiotemporal tasks, respectively. The outer ring shows video content categories derived from TV episode story summaries and standalone video descriptions. Semantically related low-frequency categories are merged for readability. Percentages are normalized independently within each ring over all QA pairs and all videos, respectively, with each TV episode counted as one video.
    }
    \label{fig:dataset_distribution}
\end{figure*}

\subsection{Harnees Details}
\label{subsec:harness}
\paragraph{Video-processing skill.}
The agent first calls \texttt{inspect\_video\_size}, which reports the sizes of its bound sources and estimates request sizes with base64 expansion and serialization overhead. When segmentation is required, \texttt{video\_process\_split} accepts only indices identifying those sources. Splitting is mandatory when request limits are exceeded and can also be required by the configured audio-transcription pathway. The FFmpeg adapter transcodes contiguous intervals and recursively bisects outputs that exceed the configured byte limit. Its manifest records source-relative temporal bounds, file sizes, and content hashes, and validation checks interval continuity and coverage. The current adapter permits an explicit exception for a final interval of at most two seconds containing no decodable video frames; any such omission is recorded in the manifest.

The separate offline \texttt{run} workflow combines Faster-Whisper large-v3 transcription with local Qwen3.6 annotation of speakers, emotions, scenes, and dialogue boundaries. ASR text, timestamps, and sentence identifiers form the transcript backbone. The annotation response must cover every supplied sentence identifier exactly once and in order, including lines explicitly marked for exclusion from dialogue outputs. The accompanying \texttt{check}, \texttt{plan}, and \texttt{validate} commands inspect readiness, preview execution, and verify generated metadata and media artifacts. These are offline CLI operations; the evaluated agent accesses the narrower inspection and splitting interfaces.

\paragraph{Visual-grounding skill.}
The agent-facing operations are \texttt{visual\_grounding\_run\_episode}, \texttt{visual\_grounding\_run\_season}, and \texttt{visual\_grounding\_run\_series}, using the configured \texttt{sam3\_1\_multiplex} model. The current interface restricts episode calls to \texttt{episode\_complex\_summary}, which returns clip counts, person and object totals, and ordered per-clip counts. Season and series calls use \texttt{scope\_person\_temporal\_summary}, which returns the total person-instance count and chronologically ordered episode or season aggregates, depending on the bound scope. The harness supplies the source set, checkpoint, and execution settings; the agent cannot substitute arbitrary paths or alter the bound visual scope.

Fresh episode inference uses trusted prepared frame inputs, while longer-scope inference reconstructs frames from the bound raw videos using the retained temporal boundaries. Counts represent distinct tracker identities within individual clips and are summed across clips, without global identity deduplication. The underlying library also supports spatial histograms using normalized image coordinates with a top-left origin, but these projections are not exposed by the current agent-facing schemas. With skill caching enabled, hybrid mode reuses provenance-matched artifacts and performs local inference on a cache miss. Cold mode requests fresh computation, whereas warm mode requires a valid cached artifact. This specialist is currently TV-scoped; standalone visual evidence is supplied through the media observer and memory retrieval.

\paragraph{Text-statistics skill.}
The offline \texttt{build} command derives dialogue-block counts, dialogue-line counts, ASR-sentence counts, and normalized speaker statistics from the configured TV annotations. The \texttt{validate} command checks source fingerprints, corpus coverage, aggregate consistency, and provenance. The agent-facing \texttt{text\_statistics\_query} exposes bounded projections through the \texttt{episode}, \texttt{season}, and \texttt{info} actions, with \texttt{info} addressing the full configured series. These deterministic CPU queries return counts, means, leading speakers, and relevant extrema without model inference.

For semantic analysis, \texttt{text\_statistics\_analyze} operates on one episode. It first retains a bounded set of non-empty dialogue lines, using uniform sampling when the episode exceeds the configured limit. Qwen3-Embedding-0.6B then matches these lines to predefined topic descriptions and ranks representative lines by embedding similarity. Qwen3-4B-Instruct-2507 receives bounded dialogue evidence and topic statistics to generate a structured summary containing themes, key events, speaker dynamics, and supporting sentence identifiers. The output is validated against the supplied identifiers, and reusable analyses follow the configured provenance-aware cache policy. These embeddings serve episode-level text analysis; the memory retriever separately uses lexical matching and temporal-neighbor expansion.

Across the three skills, the harness controls source locations, checkpoint paths, execution settings, and output budgets. The agent selects only exposed operations and validated arguments, and returned evidence excludes reference answers and private QA evidence pointers.

\paragraph{Multimodal observation.}
On a cache miss, the observation stage processes one prepared temporal unit using the actual observer's supported media interface. Native-video input is used where supported; otherwise, a compatibility adapter supplies a timestamped sequence of sampled frames. Text-only evaluated models use a separately configured multimodal observer while retaining responsibility for retrieval, tool use, and final reasoning. The observer receives neither the benchmark question nor reference answers or private evidence annotations.

For frame-based observation, the visual prompt requires the transcript field to remain empty. When the audio observer is enabled, a resident Faster-Whisper large-v3 runtime independently transcribes the source audio, using automatic language detection under the current configuration. Chunk-local transcription times are converted to source-relative times by adding the chunk's start-time offset. The transcript is then attached to the corresponding visual memory document. Missing audio tracks or the absence of detected speech produce explicitly empty transcription results. Audio and visual artifacts are cached independently, allowing valid visual observations to be reused when speech evidence is added to unchanged temporal units. The resulting evidence is accessed through memory retrieval; the agent is not given a separate transcription interface accepting filesystem paths.

\paragraph{Question-independent memory construction.}
Successful observations are normalized into documents containing a summary, transcript, on-screen text, entities, actions, events, and spatial relations, together with source identity and temporal bounds. Unavailable fields remain empty. Cache keys bind the source and temporal-unit hashes to the actual observer, observation prompt, input adapter, processing configuration, and generation settings. The audio cache additionally records the local ASR checkpoint and runtime identity. A combined audio-visual document derives its identity from both component artifacts.

When enabled, episode, season, and series manifests reference the underlying document keys and retain aggregate scope metadata. Matching observations can therefore be reused across questions and broader temporal scopes, while newly encountered units require observation. Different reasoners may share memory when the source units and actual observation configurations match. Hybrid mode reuses valid artifacts and constructs missing ones; cold mode bypasses persistent reuse, whereas warm mode requires valid cached artifacts for mandatory operations. The cache policies are summarized in next paragraph.

\paragraph{Retrieval and memory-guided reasoning.}
Retrieval is restricted to the documents bound to the current evaluation item. The initial query contains the question and any provided answer options. The retriever ranks the bounded textual representations of structured memory using term-frequency and inverse-document-frequency weights, with an additional bonus for exact query matches. It selects up to \MemoryRetrievalTopK{} seed documents and expands each seed by a \MemoryNeighborWindow{}-unit neighborhood on either side within the same source. The expanded set is deduplicated and packed under the configured record and text budgets. If the initial retrieval score falls below the configured threshold, the harness instead supplies a bounded set of evenly spaced temporal units.

The reasoner can issue further textual queries through \texttt{video\_memory\_query} within the configured execution budget. Each response retains source indices and temporal bounds and searches the same item-bound document collection. Under the cached evaluation protocol, a successful memory query is required before the final answer. Available visual-grounding and text-statistics operations provide additional evidence within their supported scopes. The final reasoning stage receives textual memory and tool results, without receiving the full raw-video history. Because observation and retrieval are bounded, persistent storage neither enlarges the native context window nor guarantees lossless retention or complete recall.

\section{Experimental Details}

\subsection{Implementation Details}
\label{app:experimental_details}

\paragraph{Evaluation scope and infrastructure.}
The evaluation comprises \StandaloneQACount{} standalone-video questions and \EvaluationSnapshotTVQACount{} television questions.
Results are grouped into visual, textual, and spatiotemporal task families, with television results further organized by episode, season, and complete-series scope.
All models listed in \cref{tab:main_answer_results} are evaluated on the complete question set under the staged \harness{}.
API inference is hosted remotely, while orchestration, media preparation, caching, retrieval, and specialist inference run on a node equipped with eight NVIDIA H100 GPUs.

\paragraph{Observation and caching.}
All staged profiles use hybrid caching, reusing artifacts whose provenance matches the requested configuration and constructing missing memory without access to benchmark questions, reference answers, or private evidence annotations.
Most multimodal profiles use the evaluated model itself as the media observer.
Qwen3.8-Max, DeepSeek-V4-Pro, and the GLM-5.3-Flash recovery profile instead use Gemini 3.1 Pro Preview to construct visual memory.
We therefore compare complete reasoner--observer systems and record both roles explicitly.
Run manifests preserve model identities, decoding and runtime settings, configuration and implementation hashes, and cache provenance to support reproducibility.

\paragraph{Answer and skill assessment.}
Answer correctness and skill use are evaluated separately.
Answer judges receive the question, candidate response, and private reference answer.
Skill-use assessment instead receives the required operations, tool traces, and execution metadata without access to the reference answer.
We report two answer-judging protocols: a single Opus-5 judge and a three-model panel comprising Kimi-K3, GLM-5.3, and Qwen3.8-Max.
The Opus-5 score follows its answer verdict without deterministic exact-match overrides.
The panel requires three valid decisions and applies a two-of-three majority rule.
Both protocols evaluate the same model responses and share the same tool-execution records.

\paragraph{Reported metrics.}
Tool invocation and execution are determined from recorded traces.
Let $N$ denote the number of evaluated questions and $B$ the number with at least one accepted tool invocation.
For each answer-judging protocol, let $C$ count questions answered correctly with all required call groups successfully executed in the prescribed order.
Let $D$ count additional correctly answered questions whose successful calls cover the required groups but do not satisfy their ordering constraints.
We report Invocation as $B/N$, Ordered as $C/B$, Order-relaxed as $(C+D)/B$, and Overall as $(C+D)/N$.
An accepted invocation may encounter a tool error; only successful calls satisfy execution requirements.
Invocation is reported once because it is identical across the two judging protocols.

The run matrix contains Gemini 3.1 Pro Preview, Gemini 3.7 Flash, GPT-5.6 Sol, Claude Opus 4.8, MiMo-V2.5, Grok-4.6, Muse-Spark-1.1, Doubao Seed 2.1 Pro (260628), GLM-5.3-Flash, Qwen3.8-Max, DeepSeek-V4-Pro, Qwen3.8-Flash, and Qwen3.7-Plus.
Exact endpoint model identifiers and resolved reasoner and observer routes are preserved in the run manifests.
Most multimodal routes use the evaluated model itself as the observer, whereas Qwen3.8-Max and DeepSeek-V4-Pro use Gemini 3.1 Pro Preview to construct visual memory; the current GLM-5.3-Flash repair profile uses the same external observer.

The local tool stack uses the SAM 3.1 multiplex checkpoint for visual grounding, Qwen3-Embedding-0.6B and Qwen3-4B-Instruct-2507 for text analysis, and Faster-Whisper 1.2.1 with the large-v3 checkpoint for audio transcription.
Text analysis uses bfloat16 inference, a batch size of 16, and an 8,192-token input limit.
Audio transcription uses float16 inference, beam size 5, voice-activity detection, and automatic language identification.
Persistent tool daemons retain model weights, and trusted GPU locks coordinate local inference across concurrent evaluation processes.

\subsection{Media preparation, caching, and retrieval}

The media adapter targets 24-second chunks with a 3.4 MB chunk-size bound and recursively splits oversized units, subject to a 20 MB request-body limit.
Claude Opus 4.8, GPT-5.6 Sol, and Grok-4.6 use timestamped JPEG frame sequences sampled at three-second intervals, capped at eight frames per chunk with a maximum image edge of 768 pixels.
Native-video routes receive bounded video inputs, and standalone frame-based observations are supplemented with local ASR.
All current staged profiles use hybrid caching, reusing only provenance-matched artifacts and constructing missing memory from the media without the question or reference answer.
Lexical retrieval selects 12 seed memory units and expands one temporal neighbor on either side within the same source.
Individual observation notes are bounded to 4,000 characters and the combined evidence summary to 64,000 characters.

\subsection{Inference budgets and execution}

The preparation and reasoning loops allow four and eight rounds, respectively, with an eight-call tool budget for each loop and a 16,000-character tool-output limit.
The default API request uses temperature 0, a 2,048-token output budget, and a 600-second timeout.
Model-specific request adapters retain the required token and reasoning controls, including an 8,192-token minimum output budget for Muse-Spark-1.1.
The evaluator selection seed is 42; local visual and text tools use seed 0, without assuming deterministic behavior from hosted APIs.
Current recovery runs use 8--48 item workers and 16--32 in-flight evaluated requests per model, with shared provider gates enforcing 60--300 requests per minute.
GPU-capable work is separately limited to four candidate workers.
Retries and resumed runs are tracked by question identifier and source/configuration provenance, and completed model answers can be rescored without rerunning inference.

\subsection{Scoring and run manifests}

Answer correctness and skill use are evaluated separately.
Private reference answers are restricted to the answer judge, while tool traces and provenance are supplied to the skill-use judge.
The current online scoring route uses \texttt{claude-opus-5}, while a separate judge-only track obtains votes from \texttt{kimi-k3}, \texttt{glm-5.3}, and \texttt{qwen3.8-max} for both dimensions.
The three-judge track requires three valid decisions before applying a two-of-three majority rule.
Missing or failed judge calls remain pending rather than being converted into negative votes, and these results are kept separate from the Opus-only scores.
Each run records the selected question identifiers, resolved model and observer routes, decoding and concurrency settings, code and configuration hashes, cache policy, and termination status.

\section{Task-Level Experimental Results and Analysis}
\label{app:full_results}

\subsection{Evaluation snapshot and reporting metric}
\label{sec:result_snapshot}

We analyze the \EvaluationSnapshotDate{}, \EvaluationSnapshotTime{} snapshot of \EvaluationSnapshotModelCount{} systems on \EvaluationSnapshotQACount{} questions, using each system's current primary answer set without selecting historical best-of runs.
This evaluation snapshot expands the earlier \TotalQACount{}-question dataset release to \EvaluationSnapshotTVQACount{} television questions and \StandaloneQACount{} standalone-video questions; it is used consistently in \cref{tab:main_answer_results} and all tables below.
The tabulated difference is confined to complete-series text and spatiotemporal group sizes, which are 36 and 24 here, compared with \TVSeriesTextCount{} and \TVSeriesSpatiotemporalCount{} in the earlier release; episode, season, and standalone group sizes are unchanged.
We use the explicit snapshot denominators below for every reported result.

For group $g$ and judging protocol $j$, let $N_g$ be the full question count, $C_{g,j}$ the number of correct answers with all required tools successfully executed in the prescribed order, and $D_{g,j}$ the number with successful required tools and a correct answer but an ordering-only violation.
The reported joint accuracy is
\begin{equation}
  A_{g,j} = 100\,\frac{C_{g,j}+D_{g,j}}{N_g}.
  \label{eq:group_joint_accuracy}
\end{equation}
Missing or failed required tools do not qualify for either numerator term.
Unfinished executions and pending or invalid judgments remain in $N_g$ without contributing to the numerator.
A dagger marks any such unresolved items, whereas a dash denotes no valid judgments and must not be read as zero accuracy.
Task-by-scope tables retain both Opus-only and three-judge results; subsequent numerical comparisons and subtype tables use the three-judge protocol unless stated otherwise.
Provisional scores describe observed joint success at this snapshot, so unequal completion and judging coverage limit comparisons of model capability.
GPT-5.6, MiMo-2.5, and Gemini-3.7 are fully executed and judged under both protocols.
Qwen3.8-Flash and Qwen3.7-Plus have 4,506 and 5,181 valid three-judge decisions, respectively; their overall joint accuracies are 64.94\% and 66.86\%.
Changes from earlier snapshots include additional executions and judgments and must not be interpreted as gains in model capability.

\subsection{Task families and temporal scopes}
\label{sec:grouped_analysis}

\paragraph{Visual rankings change across scopes.}
\Cref{tab:visual_scope_results} shows that success on episode-level visual questions does not predict the same ranking on season-level questions.
GPT-5.6 scores 99.42\% on episodes and 56.00\% on seasons, whereas Grok-4.6 scores 27.33\% and 80.00\%, respectively.
Muse-1.1 is comparatively strong on both, at 94.77\% and 76.00\%.
On standalone visual questions, the highest observed three-judge score is 54.95\% from Grok-4.6, indicating substantial remaining headroom in that group.
Complete-series visual results contain only four questions, so one answer changes the score by 25 percentage points and a 100\% result provides limited evidence of robustness.

\begin{table*}[!t]
  \caption{Visual-task joint accuracy by temporal scope in the \EvaluationSnapshotDate{} snapshot.
  Each cell reports Opus-only / three-judge accuracy (\%).
  Video denotes standalone videos; the other columns denote television scopes.
  $\dagger$ marks incomplete execution or judging, and \textemdash{} indicates no valid judgments.}
  \label{tab:visual_scope_results}
  \centering
  \begingroup
  \footnotesize
  \setlength{\tabcolsep}{3pt}
  \renewcommand{\arraystretch}{1.1}
  \begin{tabular}{@{}lrrrr@{}}
    \toprule
    Model & \shortstack{Episode\\($n=172$)} & \shortstack{Season\\($n=50$)} & \shortstack{Entire series\\($n=4$)} & \shortstack{Video\\($n=111$)} \\
    \midrule
    Gemini-3.1 & 70.93 / 70.35 & \textemdash{} / \textemdash{} & \textemdash{} / \textemdash{} & 36.04 / 34.23 \\
    GPT-5.6 & 100.00 / 99.42 & 58.00 / 56.00 & 75.00 / 75.00 & 48.65 / 45.95 \\
    Opus-4.8 & 95.35 / 94.77 & 62.00 / 60.00 & \textemdash{} / \textemdash{} & 49.55 / 52.25 \\
    MiMo-2.5 & 68.02 / 68.02 & 58.00 / 56.00 & 50.00 / 25.00 & 42.34 / 39.64 \\
    Grok-4.6 & 27.33 / 27.33 & 82.00 / 80.00 & 100.00 / 100.00 & 54.95 / 54.95 \\
    Muse-1.1 & 95.35 / 94.77 & 80.00 / 76.00 & 100.00 / 100.00 & 52.25 / 50.45 \\
    Seed-2.1 & 58.72 / 58.72 & 80.00 / 78.00 & 100.00 / 100.00 & 45.95 / 45.05 \\
    Gemini-3.7 & 36.63 / 36.63 & 82.00 / 80.00 & 100.00 / 100.00 & 48.65 / 45.95 \\
    GLM-5.3 & 62.79 / 62.79 & 78.00 / 76.00 & 100.00 / 100.00 & 47.75 / 45.95 \\
    Qwen3.8-Max & 100.00 / 99.42 & 82.00 / 80.00 & 100.00 / 100.00 & 46.85 / 46.85 \\
    DeepSeek-V4 & 99.42 / 99.42 & 82.00 / 80.00 & 100.00 / 100.00 & 40.54 / 41.44 \\
    Qwen3.8-Flash & 88.95 / 88.37 & 8.00 / 6.00 & \textemdash{} / \textemdash{} & 50.45 / 50.45 \\
    Qwen3.7-Plus & 98.26 / 97.67 & 82.00 / 80.00 & \textemdash{} / \textemdash{} & 50.45 / 45.95 \\
    \bottomrule
  \end{tabular}
  \endgroup
\end{table*}

\paragraph{Episode-level text questions offer limited separation.}
In \cref{tab:text_scope_results}, all 12 systems other than Qwen3.8-Flash score at least 98.14\% on episode-level text questions, although several entries remain provisional.
Other scopes expose larger differences: GPT-5.6 scores 96.03\% on season-level text and 72.22\% on complete-series text, while MiMo-2.5 scores 92.86\% and 91.67\%.
Standalone text accuracy also varies substantially, from a provisional 31.93\% for Seed-2.1 to 78.15\% for Gemini-3.7.
Thus, near-saturation on episode-level text questions does not imply uniformly reliable aggregation or reasoning across the other text groups.

\begin{table*}[!t]
  \caption{Text-task joint accuracy by temporal scope in the \EvaluationSnapshotDate{} snapshot.
  Each cell reports Opus-only / three-judge accuracy (\%).
  Video denotes standalone videos; the other columns denote television scopes.
  $\dagger$ marks incomplete execution or judging, and \textemdash{} indicates no valid judgments.}
  \label{tab:text_scope_results}
  \centering
  \begingroup
  \footnotesize
  \setlength{\tabcolsep}{3pt}
  \renewcommand{\arraystretch}{1.1}
  \begin{tabular}{@{}lrrrr@{}}
    \toprule
    Model & \shortstack{Episode\\($n=215$)} & \shortstack{Season\\($n=126$)} & \shortstack{Entire series\\($n=36$)} & \shortstack{Video\\($n=238$)} \\
    \midrule
    Gemini-3.1 & 99.53 / 99.53 & \textemdash{} / \textemdash{} & 50.00 / 50.00 & 59.24 / 53.36 \\
    GPT-5.6 & 100.00 / 100.00 & 96.83 / 96.03 & 75.00 / 72.22 & 71.01 / 69.33 \\
    Opus-4.8 & 98.14 / 98.14 & 85.71 / 85.71 & 50.00 / 50.00 & 72.69 / 72.69 \\
    MiMo-2.5 & 100.00 / 100.00 & 93.65 / 92.86 & 91.67 / 91.67 & 60.08 / 60.50 \\
    Grok-4.6 & 100.00 / 100.00 & 100.00 / 100.00 & 100.00 / 100.00 & 76.47 / 75.21 \\
    Muse-1.1 & 100.00 / 100.00 & 100.00 / 99.21 & 100.00 / 100.00 & 75.21 / 74.79 \\
    Seed-2.1 & 100.00 / 100.00 & 95.24 / 95.24 & 97.22 / 97.22 & 32.35 / 31.93 \\
    Gemini-3.7 & 100.00 / 100.00 & 98.41 / 98.41 & 97.22 / 97.22 & 77.31 / 78.15 \\
    GLM-5.3 & 100.00 / 100.00 & 100.00 / 100.00 & 100.00 / 100.00 & 57.56 / 55.88 \\
    Qwen3.8-Max & 100.00 / 100.00 & 98.41 / 98.41 & 100.00 / 100.00 & 68.49 / 67.65 \\
    DeepSeek-V4 & 100.00 / 100.00 & 100.00 / 100.00 & 100.00 / 100.00 & 57.98 / 57.98 \\
    Qwen3.8-Flash & 87.44 / 87.44 & 9.52 / 9.52 & 13.89 / 13.89 & 67.23 / 67.23 \\
    Qwen3.7-Plus & 100.00 / 100.00 & 100.00 / 100.00 & 97.22 / 97.22 & 66.39 / 64.71 \\
    \bottomrule
  \end{tabular}
  \endgroup
\end{table*}

\paragraph{Spatiotemporal difficulty depends on both scope and task composition.}
GPT-5.6 has complete three-judge results across all groups; its television spatiotemporal accuracy falls from 87.86\% at episode scope to 38.98\% at season scope and 12.50\% at complete-series scope (\cref{tab:spatiotemporal_scope_results}).
This decline is not explained by pending judgments, but the groups contain different reasoning operations, so it cannot be attributed solely to temporal span.
Other systems follow different profiles: Muse-1.1 scores 100.00\% on episodes and 86.44\% on seasons, while Grok-4.6 scores 91.43\% and 86.44\%.
Standalone spatiotemporal questions account for 4,039 of the \EvaluationSnapshotQACount{} snapshot items (77.46\%); consequently, a single micro-averaged benchmark score would primarily reflect this group.
The task-by-scope breakdown exposes differences that such an aggregate could obscure.

\begin{table*}[!t]
  \caption{Spatiotemporal-task joint accuracy by temporal scope in the \EvaluationSnapshotDate{} snapshot.
  Each cell reports Opus-only / three-judge accuracy (\%).
  Video denotes standalone videos; the other columns denote television scopes.
  $\dagger$ marks incomplete execution or judging, and \textemdash{} indicates no valid judgments.}
  \label{tab:spatiotemporal_scope_results}
  \centering
  \begingroup
  \footnotesize
  \setlength{\tabcolsep}{3pt}
  \renewcommand{\arraystretch}{1.1}
  \begin{tabular}{@{}lrrrr@{}}
    \toprule
    Model & \shortstack{Episode\\($n=140$)} & \shortstack{Season\\($n=59$)} & \shortstack{Entire series\\($n=24$)} & \shortstack{Video\\($n=4,039$)} \\
    \midrule
    Gemini-3.1 & 90.00 / 90.00 & - / - & 37.50 / 37.50 & 58.65 / 55.31 \\
    GPT-5.6 & 87.86 / 87.86 & 37.29 / 38.98 & 12.50 / 12.50 & 69.20 / 67.96 \\
    Opus-4.8 & 77.86 / 77.86 & 66.10 / 66.10 & 41.67 / 41.67 & 71.58 / 69.94 \\
    MiMo-2.5 & 87.86 / 87.86 & 57.63 / 55.93 & 79.17 / 70.83 & 63.88 / 61.10 \\
    Grok-4.6 & 91.43 / 91.43 & 86.44 / 86.44 & 100.00 / 100.00 & 77.57 / 74.97 \\
    Muse-1.1 & 100.00 / 100.00 & 86.44 / 86.44 & 100.00 / 100.00 & 75.96 / 74.92 \\
    Seed-2.1 & 22.14 / 22.14 & 84.75 / 84.75 & 100.00 / 100.00 & 53.92 / 52.86 \\
    Gemini-3.7 & 72.86 / 72.86 & 84.75 / 84.75 & 100.00 / 100.00 & 74.20 / 72.54 \\
    GLM-5.3 & 94.29 / 94.29 & 86.44 / 86.44 & 95.83 / 95.83 & 62.94 / 61.30 \\
    Qwen3.8-Max & 100.00 / 100.00 & 84.75 / 84.75 & 95.83 / 95.83 & 69.72 / 68.56 \\
    DeepSeek-V4 & 100.00 / 100.00 & 86.44 / 86.44 & 95.83 / 95.83 & 64.89 / 64.03 \\
    Qwen3.8-Flash & 90.71 / 90.71 & 1.69 / 1.69 & 41.67 / 41.67 & 66.82 / 66.15 \\
    Qwen3.7-Plus & 43.57 / 43.57 & 84.75 / 84.75 & 87.50 / 87.50 & 66.65 / 63.51 \\
    \bottomrule
  \end{tabular}
  \endgroup
\end{table*}

\subsection{Fine-grained television reasoning}
\label{sec:tv_subtypes}

We next examine six visual and six temporal subtypes, pooling each television subtype across its available scopes.
These tables are selected breakdowns rather than an exhaustive account of all 64 reported subtypes.
All 13 systems are included, with provisional results marked as above.

\paragraph{Arithmetic success does not generalize across aggregation operations.}
\Cref{tab:tv_visual_subtypes} localizes the reversal between GPT-5.6 and Grok-4.6.
GPT-5.6 solves all 64 person-minus-object count questions and scores 99.07\% on person-to-object count ratios, but solves only one of 17 above-mean unit-count questions (5.88\%).
Grok-4.6 scores 29.69\% and 25.93\% on the first two subtypes, but 82.35\% on above-mean counting.
Muse-1.1 scores 98.44\%, 92.59\%, and 70.59\%, respectively.
These results distinguish direct count comparisons from operations that require computing a threshold and applying it across temporal units.
Because subtype and scope are not controlled independently, the contrast identifies an operation-specific performance gap without isolating its cause.

\begin{table*}[!t]
  \caption{Television visual subtypes, pooled across temporal scopes.
  Entries are three-judge joint accuracies (\%).
  Difference and Ratio compare person instances with instances of the most frequent non-person object; Above mean counts units whose person count exceeds the mean; the remaining columns give the maximum/minimum count ratio, range divided by total count, and the share contributed by the two highest-count units.
  $\dagger$ and \textemdash{} follow \cref{tab:visual_scope_results}.}
  \label{tab:tv_visual_subtypes}
  \centering
  \begingroup
  \footnotesize
  \setlength{\tabcolsep}{4pt}
  \renewcommand{\arraystretch}{1.1}
  \begin{tabular}{@{}lrrrrrr@{}}
    \toprule
    Model & \shortstack{Difference\\($n=64$)} & \shortstack{Ratio\\($n=108$)} & \shortstack{Above mean\\($n=17$)} & \shortstack{Max/min\\($n=5$)} & \shortstack{Range/total\\($n=16$)} & \shortstack{Top two\\($n=16$)} \\
    \midrule
    Gemini-3.1 & 87.50 & 60.19 & \textemdash{} & \textemdash{} & \textemdash{} & \textemdash{} \\
    GPT-5.6 & 100.00 & 99.07 & 5.88 & 60.00 & 87.50 & 81.25 \\
    Opus-4.8 & 92.19 & 96.30 & 64.71 & 40.00 & 56.25 & 50.00 \\
    MiMo-2.5 & 78.12 & 62.04 & 41.18 & 60.00 & 75.00 & 43.75 \\
    Grok-4.6 & 29.69 & 25.93 & 82.35 & 60.00 & 87.50 & 81.25 \\
    Muse-1.1 & 98.44 & 92.59 & 70.59 & 60.00 & 87.50 & 81.25 \\
    Seed-2.1 & 65.62 & 54.63 & 82.35 & 60.00 & 87.50 & 75.00 \\
    Gemini-3.7 & 28.12 & 41.67 & 82.35 & 60.00 & 87.50 & 81.25 \\
    GLM-5.3 & 70.31 & 58.33 & 70.59 & 60.00 & 87.50 & 81.25 \\
    Qwen3.8-Max & 100.00 & 99.07 & 82.35 & 60.00 & 87.50 & 81.25 \\
    DeepSeek-V4 & 100.00 & 99.07 & 82.35 & 60.00 & 87.50 & 81.25 \\
    Qwen3.8-Flash & 92.19 & 86.11 & 11.76 & 0.00 & 6.25 & 0.00 \\
    Qwen3.7-Plus & 98.44 & 97.22 & 76.47 & 40.00 & 81.25 & 75.00 \\
    \bottomrule
  \end{tabular}
  \endgroup
\end{table*}

% \begin{samepage}
\paragraph{First-appearance reasoning and temporal aggregation separate sharply.}
GPT-5.6 solves all 64 first-appearance ordering and localization questions, yet solves only one of 16 temporal-half count comparisons (6.25\%) and one of 17 trend-reversal questions (5.88\%; \cref{tab:tv_temporal_subtypes}).
Grok-4.6 scores 98.44\%, 75.00\%, and 94.12\% on the same subtypes; Muse-1.1 scores 100.00\%, 75.00\%, and 94.12\%.
The contrast suggests that retrieving and ordering a small set of first occurrences is insufficient for reliable aggregation over an ordered sequence of counts.
It does not establish whether the failing stage is evidence retrieval, tool execution, or final computation, because joint accuracy requires all of these stages to succeed.
The five-question maximum/minimum visual subtype and other small groups should likewise be interpreted at their stated sample sizes rather than as stable rankings.
% \par
% \end{samepage}

\begin{table*}[!t]
  \caption{Television temporal subtypes over visual evidence, pooled across scopes.
  Entries are three-judge joint accuracies (\%).
  First pair compares the first appearances of two object types; First order orders first appearances and localizes their clips; Extrema gap measures ordinal distance between person-count extrema; Half diff. compares counts in the two temporal halves; Adj. change measures the largest adjacent-unit count change; Reversals counts changes in trend direction.}
  \label{tab:tv_temporal_subtypes}
  \centering
  \begingroup
  \footnotesize
  \setlength{\tabcolsep}{4pt}
  \renewcommand{\arraystretch}{1.1}
  \begin{tabular}{@{}lrrrrrr@{}}
    \toprule
    Model & \shortstack{First pair\\($n=76$)} & \shortstack{First order\\($n=64$)} & \shortstack{Extrema gap\\($n=14$)} & \shortstack{Half diff.\\($n=16$)} & \shortstack{Adj. change\\($n=16$)} & \shortstack{Reversals\\($n=17$)} \\
    \midrule
    Gemini-3.1 & 85.53 & 95.31 & \textemdash{} & \textemdash{} & \textemdash{} & \textemdash{} \\
    GPT-5.6 & 77.63 & 100.00 & 71.43 & 6.25 & 81.25 & 5.88 \\
    Opus-4.8 & 77.63 & 78.12 & 71.43 & 50.00 & 50.00 & 76.47 \\
    MiMo-2.5 & 82.89 & 93.75 & 78.57 & 62.50 & 62.50 & 23.53 \\
    Grok-4.6 & 85.53 & 98.44 & 100.00 & 75.00 & 81.25 & 94.12 \\
    Muse-1.1 & 100.00 & 100.00 & 100.00 & 75.00 & 81.25 & 94.12 \\
    Seed-2.1 & 18.42 & 26.56 & 100.00 & 68.75 & 81.25 & 94.12 \\
    Gemini-3.7 & 55.26 & 93.75 & 100.00 & 75.00 & 81.25 & 88.24 \\
    GLM-5.3 & 89.47 & 100.00 & 100.00 & 75.00 & 81.25 & 94.12 \\
    Qwen3.8-Max & 100.00 & 100.00 & 100.00 & 68.75 & 81.25 & 94.12 \\
    DeepSeek-V4 & 100.00 & 100.00 & 100.00 & 75.00 & 81.25 & 88.24 \\
    Qwen3.8-Flash & 89.47 & 92.19 & 7.14 & 0.00 & 0.00 & 0.00 \\
    Qwen3.7-Plus & 30.26 & 59.38 & 100.00 & 75.00 & 81.25 & 82.35 \\
    \bottomrule
  \end{tabular}
  \endgroup
\end{table*}

\subsection{Standalone-video spatiotemporal reasoning}
\label{sec:video_subtypes}

% \begin{samepage}
\paragraph{Distributed counting remains difficult across systems.}
The six selected video subtypes in \cref{tab:video_temporal_subtypes} reveal a common gap between event ordering and distributed counting.
GPT-5.6 scores 72.49\% on 1,374 multi-event ordering questions but 44.35\% on 372 distributed-counting questions, a 28.14-point difference with both groups fully judged.
Opus-4.8 also has fully judged distributed-counting results and scores 45.97\%; the highest observed score in that column is Muse-1.1's 47.58\%, also with complete judging.
Every system with a reported three-judge score performs worse on distributed counting than on event ordering.
This pattern is consistent with difficulty accumulating evidence across separated events, although aggregate scores alone cannot distinguish omissions, duplicate counting, and arithmetic errors.
% \par
% \end{samepage}

\paragraph{Other subtypes reveal complementary strengths.}
The highest observed trajectory-tracking score is Qwen3.8-Max's provisional 66.46\%, while Muse-1.1 has the highest observed state-change score at a provisional 73.80\%.
Grok-4.6 and Muse-1.1 both score 78.80\% on fully judged cross-modal reasoning, and Grok-4.6 scores 81.58\% on fully judged cross-modal binding.
These profiles support evaluating evidence accumulation, temporal order, trajectories, and modality binding separately, rather than treating spatiotemporal reasoning as a single uniform capability.

\paragraph{Judge choice and coverage bound the conclusions.}
The paired task-by-scope scores also expose sensitivity to the evaluation protocol.
For MiMo-2.5 on complete-series spatiotemporal questions, Opus-only and three-judge scores are 79.17\% and 70.83\%, with complete judging under both protocols.
GPT-5.6 also has complete judging under both protocols, but its standalone spatiotemporal score changes from 69.20\% to 67.96\%.
These aggregate differences do not quantify inter-judge agreement, which requires a separate comparison of matched item-level votes.
The results characterize the evaluated reasoner--observer--tool systems under the stated metric; without matched standalone-model runs or component ablations, they do not measure a causal gain from the harness or identify a unique failure mechanism.

\section{Discussion}

\subsection{Limitations}
\label{app:limitations}
Our findings should be interpreted within the benchmark's current coverage and evaluation setting.
The serialized collection covers two sitcoms, and the task distribution emphasizes spatiotemporal reasoning over video-derived evidence.
Automated annotation and model-based judging introduce uncertainty, while performance depends on observation quality, retrieval budgets, and tool execution.
The reported scores therefore characterize the evaluated reasoner--observer--tool systems as a whole.

\subsection{Future Work}
\label{app:future_work}
Future work will explore broader domains and continuous streams to evaluate memory updates, entity persistence, and reasoning across temporal scales.
Matched evaluations with controlled observation inputs and inference budgets can clarify the contributions of memory representations, retrieval policies, and specialist tools.
Promising directions include quantity-aware memory with explicit temporal provenance, uncertainty-guided retrieval, and source-level human verification to strengthen evaluation reliability.

\end{document}